\documentclass[letterpaper, 10 pt, journal, twoside]{ieeetran} 
\usepackage{amsmath,amsfonts}
\usepackage{algorithmic}
\usepackage{algorithm}
\usepackage{array}
\usepackage[caption=false,font=normalsize,labelfont=sf,textfont=sf]{subfig}
\usepackage{textcomp}
\usepackage{stfloats}
\usepackage{url}
\usepackage{verbatim}
\usepackage{graphicx}
\usepackage{cite}
\usepackage[breaklinks,hidelinks]{hyperref}
\usepackage{cleveref}
\usepackage{siunitx}
\usepackage[nolist,nohyperlinks]{acronym}
\usepackage{booktabs}
\usepackage{makecell}
\usepackage{multirow}
\acrodef{rl}[RL]{Reinforcement Learning}
\acrodef{mlp}[MLP]{Multi-Layer Perceptron}
\acrodef{lstm}[LSTM]{Long Short-Term Memory}
\acrodef{fov}[FOV]{Field-of-View}
\acrodef{appo}[APPO]{Asynchronous Proximal Policy Optimization}
\acrodef{rnn}[RNN]{Recurrent Neural Network}
\acrodef{gru}[GRU]{Gated Recurrent Unit}
\acrodef{wbt}[WBT]{Water Ballast Tank}
\acrodef{gvi}[GVI]{General Visual Inspection}
\acrodef{pomdp}[POMDP]{Partially Observable Markov Decision Process}
\acrodef{svs}[SVS]{Spatial Visit Score}

\begin{document}
\title{Semantically-driven Deep Reinforcement Learning\\ for Inspection Path Planning}

\author{Grzegorz Malczyk$^{*}$, Mihir Kulkarni and Kostas Alexis
\thanks{Manuscript received: March, 18, 2025; Accepted May, 18, 2025.}
\thanks{This paper was recommended for publication by Editor Giuseppe Loianno upon evaluation of the Associate Editor and Reviewers' comments. This work was supported by the Research Council of Norway under Award NO-338694 and the Horizon Europe Grant Agreement No. 101119774. The authors thank the SINTEF Multiphase Flow Laboratory for providing access to the testing site for experimental evaluations.} 
\thanks{The authors are with the Department of Engineering Cybernetics, Norwegian University of Science and Technology (NTNU), Norway.}
\thanks{$^*$Corresponding author. {E-mail: \tt\footnotesize grzegorz.malczyk@ntnu.no}}
\thanks{Digital Object Identifier (DOI): see top of this page.}

}
\markboth{IEEE Robotics and Automation Letters. Preprint Version. Accepted May, 2025}
{Malczyk \MakeLowercase{\textit{et al.}}: Semantically-driven Deep Reinforcement Learning for Inspection Path Planning}  
\maketitle

\begin{abstract}
This paper introduces a novel semantics-aware inspection planning policy derived through deep reinforcement learning. Reflecting the fact that within autonomous informative path planning missions in unknown environments, it is often only a sparse set of objects of interest that need to be inspected, the method contributes an end-to-end policy that simultaneously performs semantic object visual inspection combined with collision-free navigation. Assuming access only to the instantaneous depth map, the associated segmentation image, the ego-centric local occupancy, and the history of past positions in the robot's neighborhood, the method demonstrates robust generalizability and successful crossing of the sim2real gap. Beyond simulations and extensive comparison studies, the approach is verified in experimental evaluations onboard a flying robot deployed in novel environments with previously unseen semantics and overall geometric configurations.

%

\end{abstract}


\vspace{-2ex}
\section{Introduction}
\begin{figure}
    \centering
    \includegraphics[width=0.98\columnwidth]{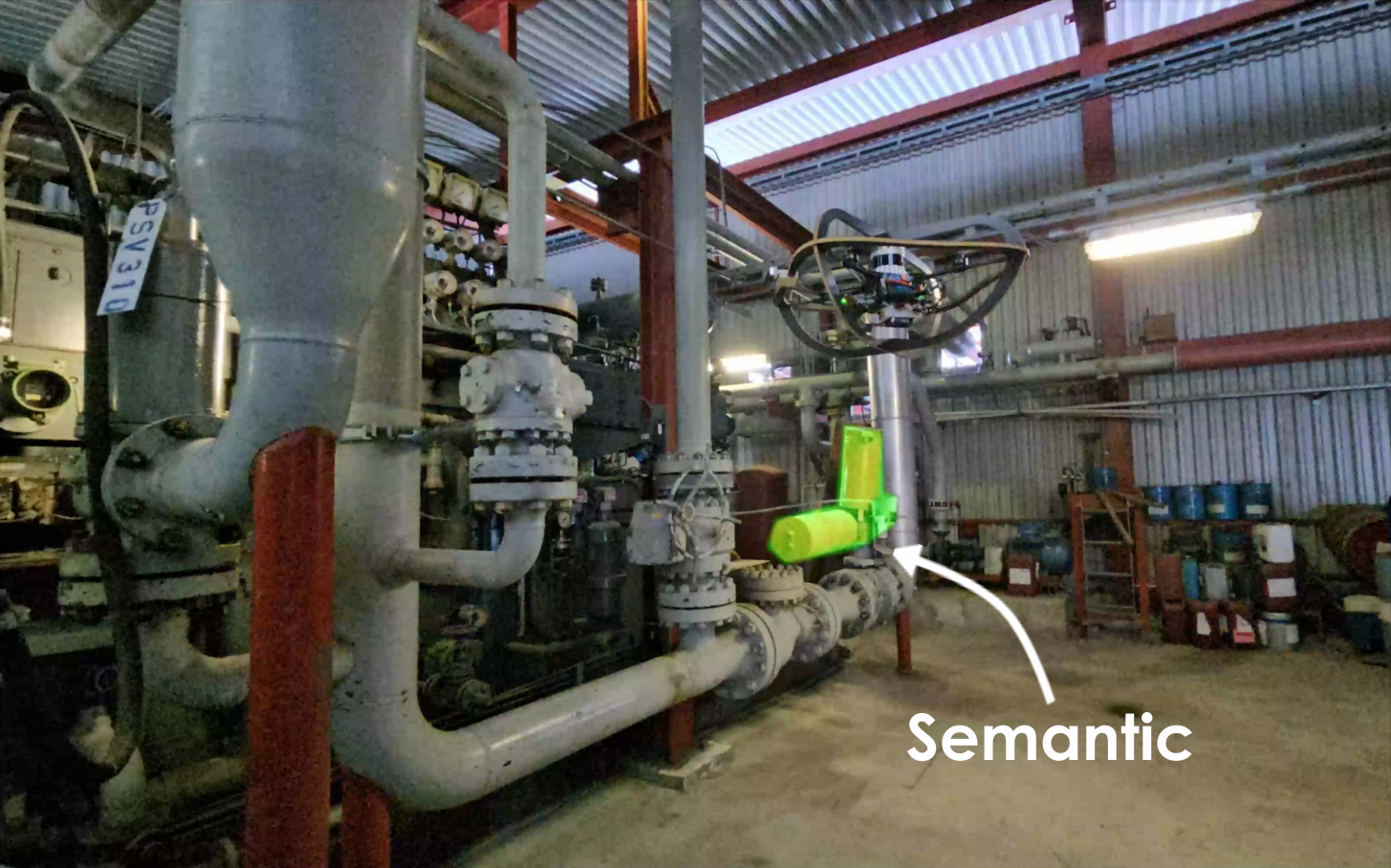}
    \vspace{-2ex}
    \caption{A flying robot conducting an inspection in an industrial environment.}
    \label{fig:title_image}
    \vspace{-4ex}
\end{figure}

\IEEEPARstart{A}{utonomous} robotic inspection has emerged as a critical capability for tasks ranging from industrial facility monitoring to disaster response. Despite significant advances in aerial robotics, a fundamental challenge persists: how to efficiently inspect only what matters in complex, unknown environments. Unlike traditional exploration, where complete coverage is the goal, real-world inspection missions often require attention to specific objects or surfaces of interest - what we term ``semantics'' - while safely navigating through the environment. For instance, in an industrial facility, such as the one illustrated in~\Cref{fig:title_image}, within a space filled with diverse structures, the robotic system should discern and prioritize critical inspection targets, focusing computational and navigational resources on areas of substantive interest.

Traditional exploration and coverage planning algorithms, while effective for comprehensive mapping, prove inefficient for such targeted inspection tasks~\cite{mccammon2021topological,dharmadhikari2021hypergame, zhou2021fuel,moon2022tigris,dharmadhikari2023autonomous}. These approaches typically either exhaustively cover all surfaces without discriminating their relevance or require extensive environment-specific tuning to achieve acceptable performance. Often, they fail to prioritize regions of higher importance, resulting in wasted resources and longer mission times~\cite{galceran2013survey}. Furthermore, such methods typically struggle with dynamic adaptation to unexpected environmental changes during mission execution, as they tend to rely on predefined models or static assumptions about the scene.



Among the highest-performing methods in informative path planning are those based on sampling techniques, which efficiently explore environments but often lack semantic awareness, potentially leading to suboptimal allocation of inspection effort~\cite{schmid2020efficient}. To address this, semantically-aware approaches integrate detailed mapping and planning modules to prioritize regions of interest, enhancing inspection efficiency~\cite{wang2019semantic,dharmadhikari2023semantics,lu2024semantics}. However, these methods often require complex environment representations, introducing computational overhead, particularly in large-scale applications, and typically rely on prior semantic knowledge, limiting adaptability to new scenarios.


Recent advances in \ac{rl} have shown promise for informative path planning by learning to optimize actions for maximum information gain~\cite{chen2021zero}. These approaches leverage various state representations, including agent pose~\cite{chen2022learning}, occupancy grids~\cite{lodel2022look}, sensor history~\cite{cao2023catnipp}, and action-observation histories~\cite{choudhury2020adaptive}. However, relatively few studies have incorporated semantic information into learning-based path planning objectives~\cite{blum2019active}, despite its demonstrated benefits in navigation tasks~\cite{santos2022deep,kastner2022enhancing,chaplot2020object,kulkarni2023semantically}. Moreover, \ac{rl}-based methods face several challenges, including data-intensive training requirements, sensitivity to reward design, and the sim2real gap, all of which limit their real-world applicability~\cite{popovic2024robotic}. While some works suggest that integrating semantic awareness could enhance robustness and adaptability~\cite{schmid2022sc}, its full potential in \ac{rl}-based path planning remains underexplored.



\vspace{-2ex}
\subsection{Contributions}

Motivated by the above, we contribute a novel semantics-aware information planning policy based on deep \ac{rl} that learns to visually inspect the objects of interest while enabling collision-free flight. The policy operates through a straightforward reward function design while relying exclusively on local information available from onboard sensors, namely a)~odometry, b)~instantaneous depth measurements, c)~semantic segmentation masks, and d)~dynamically constructed ego-centric local occupancy and position-based maps. Critically, the method does not assume prior map knowledge or consistent long-term mapping, nor is it limited to predefined sets of semantic classes, their shape, size, or location. The inspection is performed at any desired resolution/distance to the object up to what is possible through collision-free configurations. To demonstrate its performance, the method is verified both in extensive simulations as well as in experimental evaluations onboard a flying robot. To the authors' best knowledge, the presented work represents the first semantics-aware end-to-end \ac{rl} inspection planning framework. To support reproducibility, the method is open-sourced in \url{https://github.com/ntnu-arl/semantic-RL-inspection}.
\subsection{Outline}
In the remainder of this paper, \Cref{sec:problem_formulation} presents the problem formulation, followed by the description of the proposed method in \Cref{sec:method}. Evaluation studies are presented in the \Cref{sec:results}. Finally, conclusions are drawn in \Cref{sec:conclusion}.
\section{Problem formulation}
\label{sec:problem_formulation}
Consider a robot in an unknown local 3D environment that has to find and inspect a given set~$\Lambda$ of semantic target objects~$\mathbb{S}_i$, represented as triangle meshes with shapes and locations that are not known a priori. The semantic inspection planning problem, as considered in this work, is that of incrementally deriving a collision-free coverage path~$\mathcal{P}_i$ for the surfaces of~$\mathbb{S}_i$ assuming access only to a)~the estimate of the robot's state $\mathbf{s}_t$, b)~the current depth image observation $\mathbf{D}_t$, c)~an associated segmentation mask $\mathbf{S}_t$, d)~an ego-centric local occupancy map $\mathbf{m}^o_t$ aligned with the robot's orientation, and e)~a history of robot’s 3D position trajectory $\mathbf{T}_{1:t}$,
all obtained from onboard measurements at time instant $t$.
The estimated robot state is defined as:

\vspace{-5mm}
\begin{align}
    \mathbf{s}_t = [ \mathbf{p}_t^\top, \mathbf{q}_t^\top, \mathbf{v}_t^\top, \boldsymbol{\omega}_t^\top ]^\top,
    \label{eq:state}
\end{align}
which consists of its 3D position $\mathbf{p}_t$, orientation in a 4D vector form $\mathbf{q}_t$, 3D linear velocity $\mathbf{v}_t$ and 3D angular velocity $\boldsymbol{\omega}_t$, all expressed in the inertial reference frame $\mathcal{I}$. A mask $\mathbf{S}_t$ is a binary image where the pixels corresponding to the semantic of interest are set to $1$.
$\mathbf{D}_t$ is the depth image. Both images can be obtained from an onboard RGB-D camera device as in the studies of this work. 
The local ego-centric occupancy map is built based on the range sensor readings while the trajectory $\mathbf{T}_{1:t}$ stores all the robot's 3D positions until time $t$.
Given an unknown 3D object of interest in the environment, the goal is to iteratively find an optimized action vector:
\vspace{-1mm}
\begin{align}
    \mathbf{u}_t = [{\mathbf{v}_t^{r}}^\top, {\omega}_{t,z}^{r}]^\top,
\label{eq:action}
\end{align}
involving the commanded robot reference linear velocities $\mathbf{v}_t^{r}\in \mathbb{R}^3$ and the yaw rate ${\omega}_{t,z}^{r}$ (both expressed in the vehicle frame $\mathcal{V}$) such that the robot a) localizes and inspects the semantic of interest leading to maximum coverage of its surface, while b) flying safely by avoiding collisions despite the presence of nearby obstacles. Finally, the action vector $\mathbf{u}_t$ is fed into a low-level controller of the flying vehicle as a commanded value. Although the problem is formulated for inspecting one object at a time, the approach generalizes to scenes with multiple semantics, as shown in \Cref{sec:results}.
\section{Method}
\label{sec:method}
We solve the semantic inspection problem by formulating it as a reinforcement learning task. In this work, the state space~$\mathcal{S}$ is defined as the set of all possible agent and environment states with $\mathfrak{s}_t\in\mathcal{S}$ at discrete time $t$. The action space~$\mathcal{A}$ consists of all possible actions that the robot could take with~$\mathbf{a}_t\in\mathcal{A}$. The observation space~$\mathcal{O}$ corresponds to the set of all possible sensor readings and associated interpretations the robot could have about the surroundings with $\mathfrak{o}_t\in\mathcal{O}$. Following, we define how we construct each of these parts of the \ac{rl} problem for inspection path planning.

\begin{figure*}[t]
    \centering
    \includegraphics[width=0.98\textwidth]{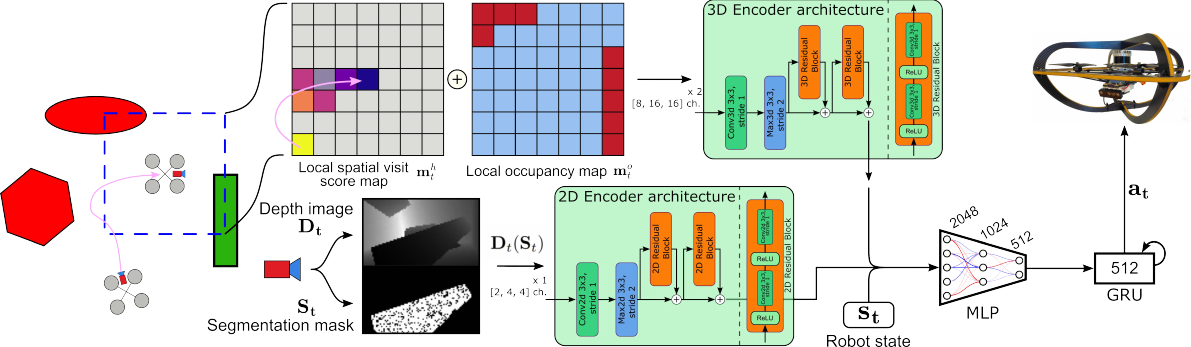}
    \vspace{-2ex}
    \caption{
    The proposed deep \ac{rl} network for semantics-driven inspection. For a robot navigating an environment a target semantic object and obstacles, the local region from which occupancy and spatial visit score maps are derived is outlined with a dashed line. The two local maps are processed by the 3D encoder, while a semantically masked depth image is fed into the 2D encoder. The resulting latent representations, combined with the agent's state, are passed through an MLP and then a GRU block.
    }
    \label{fig:network_architecture}
    \vspace{-3ex}
\end{figure*}
\subsection{Ego-centric Occupancy Map}
\label{sec:occupancy}
The occupancy map enables safe navigation by providing a representation of the environment's spatial structure. When moving in the environment, the robot builds a global but possibly drifting occupancy map $\mathbf{M}^o_t$ from range measurements only. At every time step $t$, the robot makes an observation, given as a point cloud from the sensor, and updates its belief about the target map. The robot’s belief about the map $\mathbf{M}^o$ is a discretization of the world in grid cells of pre-defined resolution $r_V$. Each of the cells has an associated status of being occupied, free or unknown. 

The proposed \ac{rl} framework operates on the local occupancy map $\mathbf{m}^o$ for better generalization across different environments and computational efficiency. To obtain $\mathbf{m}^o$, we query corresponding cell values from $\mathbf{M}^o$. The local occupancy map is then given as a $n\times n\times n$ grid matrix centered at the robot’s position, aligned with its orientation~\cite{chen2019learning,pfeiffer2018data}, and expressed in the body frame $\mathcal{B}$. Each cell's status corresponds to the associated voxel value in the global map, providing the \ac{rl} agent with local obstacle information to apply collision-free actions. Drift in the global map $\mathbf{M}^o$ may be  of limited importance,
given the highly local nature of $\mathbf{m}^o$, which represents a cubic volume with a side length of~\SI{2}{\meter}, centered around the agent. A horizontal slice of the 3D ego-centric occupancy grid is visualized in \Cref{fig:network_architecture}.

\subsection{\ac{svs} Map}
\label{sec:entropy}

The robot's understanding of its 3D position history $\mathbf{T}_{1:t}$ in the environment is represented by a \ac{svs} map, denoted as $\mathbf{m}^h$. This map captures the variability of visit frequencies across different regions of the local environment, assigning a measure of visit dispersion to each cell~\cite{niroui2019deep,zhu2018deep}. 
For generalization and efficient training, the robot is provided with a 3D ego-centric local \ac{svs} map, expressed in frame $\mathcal{B}$ and constructed similarly to the ego-centric local occupancy map. We define this local map $\mathbf{m}^h$ as a cube of size $n\times n\times n$ with edge size~$r_V$.
Each cell in the local map is assigned a spatial visit score, based on the robot's trajectory $\mathbf{T}_{1:t}$ and the total number of visits across the local map $N_t$. This score is then calculated using the Shannon entropy formula:
\begin{align} 
    \mathbf{m}^h_{i,j,k} = - p_{i,j,k} \log{p_{i,j,k}}, 
\end{align} 
where $p_{i,j,k}$ represents the normalized number of visits to cell $(i,j,k)$ within the local map region.


\subsection{Inspection Policy Learning}
We employ a deep reinforcement learning framework to train, in an end-to-end manner, policies to inspect a semantic of interest~$\mathbb{S}_i$ while navigating within cluttered environments. We formulate the problem as a \ac{pomdp} with the observation vector $\mathfrak{o}_t$,
agent state denoted by $\mathfrak{s}_t$ and action $\mathbf{a}_{t}$ at a discrete time $t$. 

\paragraph{Observations and Actions Formulation} At each step, the agent observes the environment given by the following observation vector:
\vspace{-1mm}
\begin{align}
    \mathfrak{o}_t = [\mathbf{s}_{t}, \mathbf{a}_{t-1}, \mathbf{D}_t(\mathbf{S}_t), \mathbf{m}^o_t, \mathbf{m}^h_t],
\label{eq:observation}
\end{align}
where $\mathbf{D}_t(\mathbf{S}_t)$ is the masked depth image based on the semantic mask. We calculate it as $\mathbf{D}_t \circ \mathbf{S}_t$, where the $\circ$ operator denotes the Hadamard product. This observation choice allows to generalize across various environments while it also prevents overfitting to the particular shape of the inspected objects. An action of the agent takes the form $\{\mathbf{a}_t \in \mathbb{R}^4 | -1 \leq \mathbf{a_{t}}_i \leq 1\}$ and is applied to the environment to obtain a new state at the next time step $t + 1$ with the transition probability $Pr(\mathfrak{s}_{t+1}|\mathfrak{s}_t, \mathbf{a}_t)$. The action $\mathbf{a}_t$ is converted to velocity and yaw-rate command as:
\vspace{-1mm}
\begin{align}
    \mathbf{u}_t = \mathbf{a}_t \circ \mathbf{w}_{max},
\end{align}
where $\mathbf{w}_{max} \in \mathbb{R}^4$ defines the maximum velocity for each component, as formulated in~\Cref{eq:action}. Thus, we allow the agent to fully utilize the range of possible motions and efficiently explore the environment without constraining the commands to lie within the \ac{fov} of the range sensor. Likewise, this representation implies that the proposed \ac{rl} inspection planning framework is tailored to exploit agile maneuvering as necessary without being limited to the simplification of offering a list of waypoints that neglect the vehicle's dynamics as commonly found in path planning~\cite{bircher2016receding,bircher2018receding,englot2012sampling}. 

\paragraph{Reward Design} For each state transition, a reward is provided to the agent in the form of $\mathcal{R}(\mathfrak{s}_t, \mathbf{a}_t)$. Given observations and previous actions, a policy that yields the current actions $\mathbf{a}_t = \pi(\mathfrak{o}_{t}, \mathfrak{b}_{t})$ is learned by the agent, where $\mathfrak{b}_{t}$ is the belief over the true states of the environment. The learning process aims to maximize the sum of rewards over an episode. The given semantic is assumed to be a mesh composed of $N_f$ faces. The reward function for the state transitions is defined as:

\vspace{-5mm}
\begin{align}
    \mathcal{R}(\mathfrak{s}_t, \mathbf{a}_t) = f_t + v_t + p_t,
\end{align}
where each term serves a distinct purpose. The face-mesh reward $f_t$ drives the semantic surface coverage, the semantic-search reward $v_t$ encourages the exploration of semantic object predominantly when it has not already been observed by the agent, and the collision penalty $p_t$ prevents from getting too close to obstacles. This minimal yet effective formulation ensures an interpretable reward structure.

\begin{figure}
    \centering
    \includegraphics[width=0.98\columnwidth]{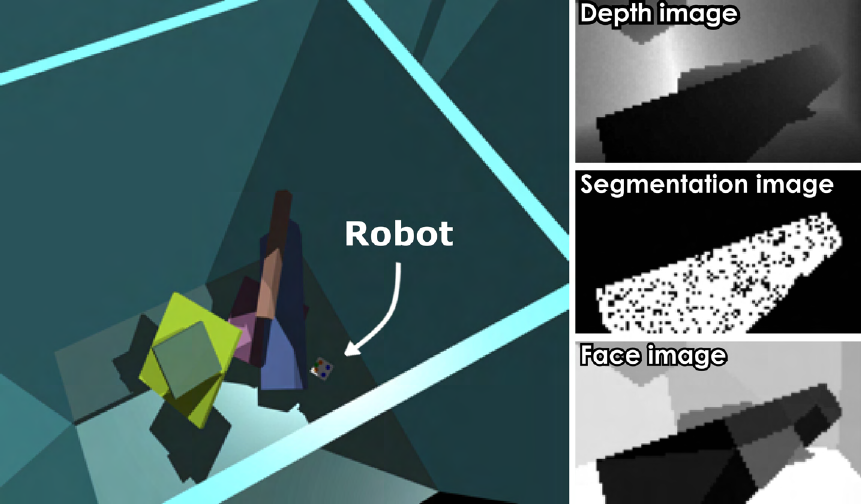}
    \vspace{-2ex}
    \caption{\textit{Left:} Inspection mission in the Aerial Gym Simulator. The environment consists of obstacles and a semantic of interest presented as a blue cuboid in this scenario. \textit{Right top-down:} Images utilized during the training.}
    \vspace{-3ex}
    \label{fig:aerial_gym}
\end{figure}

The face-mesh reward $f_t \in [0, 1]$ is the fraction of the inspected faces of a semantic of interest seen with the camera sensor until time $t$, weighted by observation distance. At each time-step $t$, a privileged image corresponding to the depth image is obtained from the simulator that annotates each pixel with the index of the mesh-face~(\Cref{fig:aerial_gym}).
For each face, we compute the mean depth distance $d$ in the current camera frame and compare it with the previously stored value for that face $d_f$. 
We update the respective face depth value $d_f$ if $d$ is closer to the desired inspection distance $d_{ref}$. Then, the inspection score $n_f$ for each face is:

\vspace{-3mm}

\small
\begin{equation}
    n_f = \left\{ 
  \begin{array}{ c l }
    \mathbf{\alpha} \mathrm{e}^{(-\beta (d - d_{ref})^2)} & \quad \textrm{if } |d - d_{ref}| < |d_f - d_{ref}|  \\
    0                 & \quad \textrm{otherwise,} \end{array} \right.
    \label{eq:reward_function}
\end{equation}
\vspace{-3mm}
\begin{align}
    & d_f = \left\{ 
  \begin{array}{ c l }
    d   & \quad \textrm{if } n_f > 0  \\
    d_f & \quad \textrm{otherwise,} \end{array} \right.
\end{align}
\normalsize
where $d$ is the current mean distance between the camera and the face. The parameters $\alpha,\beta>0$ represent tuning parameters. Accordingly, the reward term $f_t$ represents the accumulated sum of the fraction of inspected faces:
\vspace{-1mm}
\begin{equation}
    f_t = \sum_{N_f} n_f,
\end{equation}
which is provided to the agent at every action step. Thus, the value $f_t$ implicitly encodes the desired resolution on which the inspected object's faces should be visually monitored.
Additionally, we assume that a specific face of the semantic object has been seen if it falls within an image area, which we define as a focus area. This image region is defined as half of the image's dimensions, centered around the visual center of the image.

The semantic-search reward $v_t$ is given as
\vspace{-1mm}
\begin{align}
    v_t = \gamma \mathrm{e}^{-\delta N_t},
\end{align}
where $N_t$ is the number of visits within the local map determined by the history of robot’s trajectory and the volume of the local map. Additionally, the parameters $\gamma,\delta>0$ represent tuning parameters. 

The collision penalty $p_t$ is set to~$-1$ if any occupied voxels in $\mathbf{m}^o_t$ are within $d_{coll}$ of the agent and $0$ otherwise. This collision distance reflects the robot's physical dimensions.

\subsection{Implementation}

\label{sec:implementation}
We utilize the \ac{appo} algorithm from Sample Factory \cite{petrenko2020sample} to train a deep neural network policy to navigate a robot and perform the inspection task in a collision-free manner. We extend a 2D model and the hyperparameters proposed in~\cite{espeholt2018impala} to the 3D ResNet model to encode spatial information from the local 3D map inputs $\mathbf{m}^o$ and $\mathbf{m}^h$, as shown in \Cref{fig:network_architecture}.
Ego-centric maps of size $n\times n\times n$, where $n=21$ are discretized using a grid size of $r_V=$~\SI{0.1}{m} and the collision distance $d_{coll}$ is set to \SI{0.3}{\meter}. Thus, compressed latent representations of spatial features in the local obstacle grid and the \ac{svs} map are learned. These local occupancy features enable efficient learning of collision avoidance with objects of different shapes while the \ac{svs} map enhances the agent’s environment exploration capabilities.

Next, we stack both image inputs to encode the depth and segmentation information of the target object, $\mathbf{D}_t(\mathbf{S}_t)$, from the 2D images using the 2D ResNet encoder proposed in~\cite{espeholt2018impala}. For generalizability across different environments and to enhance the sim2real transfer while keeping the network architecture small, before encoding the 2D input we down-sample the image to a size of $96 \times 54$ pixels and obtain a single depth image masked out by the semantic image. 

Finally, the policy exploits the 2D and 3D spatial features to find actions that guide the robot around nearby obstacles and perform local, semantic-oriented search in the environment followed by the task of visual inspection of the target object. The two latent feature vectors are concatenated with the current robot state $\mathbf{s}_t$ and fed into a \ac{mlp} network consisting of three fully connected layers of size $2048$, $1024$ and $512$ neurons each, with an ELU activation layer, followed by a \ac{gru} block with a hidden layer size of $512$. The incorporation of the memory unit inside the policy improves the performance during the inspection phase. Given an observation vector $\mathfrak{o}_t$, the policy outputs a 4-dimensional action command $\mathbf{a}_t$. With $\mathbf{w}_{max}$ set to \SI{1}{\meter/\s} for linear velocity and \SI{1}{\radian/\s} for yaw-rate, the command is sent to the low-level velocity controller onboard the robot, as shown in \Cref{fig:network_architecture}.

\subsection{Training Environment}
For training, we use the Aerial Gym Simulator \cite{kulkarni2025aerialgym_journal}, which provides the environment and the interfaces to train the deep \ac{rl} policy to navigate within various environments. The simulator offers the capabilities for massively parallelized simulation of aerial robots with exteroceptive sensors. The simulated flying platform is equipped with an RGB-D camera with horizontal and vertical \ac{fov} of $\{87,58\}$ \SI{}{\deg}, a LiDAR with \ac{fov} of $\{360,90\}$ \SI{}{\deg}, and the quadrotor velocity controller in \cite{lee2010control}. We generate the environments within the simulator consisting of room-like scenes containing static obstacles of primitive shapes and different sizes. The utilization of primitive geometric objects (e.g., boxes, cylinders or spheres) relates to the goal of generalizability without considering any specific shape of semantics. The objects are positioned and oriented in a randomized manner, and a unique semantic label is assigned to each of them. We run the learning model with only nine obstacles and one semantic of interest with $N_f=60$, allowing us to train the policy in approximately six hours on an NVIDIA RTX A6000. The images~$\mathbf{D}_t,~\mathbf{S}_t$ are derived from the data of depth and segmentation cameras onboard the simulated robot. Global \ac{svs} and occupancy maps are created as the robot explores the environment, defined in the world frame with the cell size of \SI{0.1}{\m}, and used to extract the local ego-centric maps as inputs to the network as described in \Cref{sec:occupancy} and \Cref{sec:entropy}. 
Each episode involved a room-shaped environment with dimensions $L \times W \times H$ within the set $[4, 20] \times [4, 20], \times [2, 8]$~\SI{}{\meter}. An example training environment is presented in \Cref{fig:aerial_gym}. The desired inspection distance $d_{ref} \in [0.5, 2.5]$~\SI{}{\meter} is used as a hyper-parameter and is chosen according to the configuration of the deployed onboard camera. The length of the episode is defined as~\SI{90}{\s} considering the size of the environments and the size of the objects to be inspected. 

To robustify the network performance against real-world uncertainty, random disturbance wrenches are applied to the simulated platform, sampled from a normal distribution. The state observation from~\Cref{eq:state} is also perturbed by random noise sampled from a uniform distribution to simulate the uncertainty from real-world sensor measurements of the agent state. Moreover, we limit the image capture rate to~\SI{10}{Hz}, while the physics simulation occurs at~\SI{100}{Hz}. Finally, we add Gaussian noise to the simulated images by sampling from a distribution with the standard deviation linearly dependent on the depth value of the pixel. 


\begin{table}[t]
\centering
\caption{Evaluation of the trained policy against environments for different levels of complexity.}
\vspace{-2ex}
\label{tab:curriculum}
\begin{tabular}{@{}cccccc@{}}
\toprule
\multicolumn{1}{l}{\# obstacles}  & 0        & 4              & 9             & 14                & 19      \\ \midrule
Avg. Coverage                     & 92.1\%   & 82.6\%         & 74.6\%        & 64.2\%            & 58.1\%     \\
Crash                             & \textbf{0.0\%}    & \textbf{0.3}\%          & \textbf{1.2}\%         & \textbf{1.5}\%             & \textbf{1.3\%}     \\
Timeout                           & 100.0\%  & 99.7\%         & 98.8\%        & 98.5\%            & 98.7\%     \\ \cmidrule(l){1-6}     
\end{tabular}
\vspace{-3ex}
\end{table}

\begin{table}[t]
\centering
\caption{Impact of the local \ac{svs} map on semantic inspection in an environment with nine obstacles.}
\vspace{-2ex}
\label{tab:entropy}
\begin{tabular}{@{}cccc@{}}
\toprule
\multicolumn{1}{l}{}        & $\mathbf{m}^o$       & $\mathbf{m}^o$ \& $\mathbf{m}^h$\\ \midrule
Avg. Coverage               & \textbf{52.3}\%       & \textbf{74.6\%}             \\
Crash                       & 1.0\%        & 1.2\%              \\
Timeout                     & 99.0\%       & 98.8\%             \\ \cmidrule(l){1-3}     
\end{tabular}
\vspace{-4ex}
\end{table}

\begin{table}[t]
\centering
\caption{Impact of the \ac{gru} unit on semantic inspection performance in an environment with nine obstacles.}
\vspace{-2ex}
\label{tab:gru}
\begin{tabular}{@{}ccccc@{}}
\toprule
\multicolumn{1}{l}{}        & no \ac{gru}    & \ac{gru} 128       & \ac{gru} 512    \\ \midrule
Avg. Coverage               & \textbf{59.8}\%          & \textbf{63.6}\%              & \textbf{74.6}\%             \\
Crash                       & 3.6\%  & 1.1\%               & 1.2\%              \\
Timeout                     & 96.4\%          & 99.1\%              & 98.8\%             \\ \cmidrule(l){1-4}     
\end{tabular}
\vspace{-4ex}
\end{table}

\section{Results}
\label{sec:results}
In this section, we demonstrate the effectiveness of our semantics-aware inspection planning \ac{rl} policy through a comprehensive evaluation spanning both high-fidelity simulations and real-world robotic deployments. Our experiments validate the method's performance across diverse environmental configurations, revealing its robust generalization capabilities and successful bridging of the sim2real gap.
\subsection{Ablation Studies}

We conducted an ablation study to evaluate the planner in a room-like environment with dimensions of $10\times10\times 4$~\SI{}{\meter}, featuring a semantic object and obstacles of primitive shapes.
To assess the training performance, we set up a curriculum that logs runs of the robots measuring semantic surface coverage, crash incidents, and episode timeouts. 
Due to environmental randomization, in some situations, the feasible surface to inspect is below~$100$\%  physically blocking access to certain areas or causing occlusions of the target object.
We define instances where the agent is in contact with any other object as crashes. A timeout occurs when the robot has remained collision-free till the end of the episode. 

The results, presented in \Cref{tab:curriculum}, show the planner's performance as the number of obstacles increases while inspecting a single semantic object. The simulation was run across 512 environments for 6 episodes each. We report the average semantic surface coverage, as well as crash and timeout rates. Thanks to the purely local observations, the model generalizes well for more cluttered environments, and the planner successfully performs the visual inspection task in more complex environments while avoiding collisions. The reduction in average semantic surface coverage is attributed to fewer feasible viewpoints for inspection in more cluttered environments.

In \Cref{tab:entropy}, we analyze the effect of the local \ac{svs} map on inspection performance. We retrained the policy described in \Cref{fig:network_architecture}, excluding the \ac{svs} map as an observation for the network. The modified policy was tested in a room-like environment with nine obstacles and one semantic object to inspect. The \ac{svs} map allows the agent to efficiently explore the local environment, discover new feasible viewpoints and increase semantic surface coverage to $74.6\%$. 

Finally, we evaluate the impact of the memory unit within the \ac{rl} pipeline. As described in \Cref{sec:method}, we incorporate a \ac{gru} unit as the \ac{rnn}. In \Cref{tab:gru}, we show that the inclusion of the \ac{gru} block enhances the agent's ability to capture long-term dependencies during the inspection task. This improvement stems from the \ac{gru}’s ability to handle variable-length dependencies, which is particularly beneficial in inspection tasks where initial exploration plays a crucial role. Without a memory unit, the network struggles to learn the inspection task effectively, likely due to the nature of the reward signal in \Cref{eq:reward_function} guiding the learning process.

\subsection{Simulation Studies}
\begin{figure}[t]
    \centering
    \includegraphics[width=0.98\columnwidth]{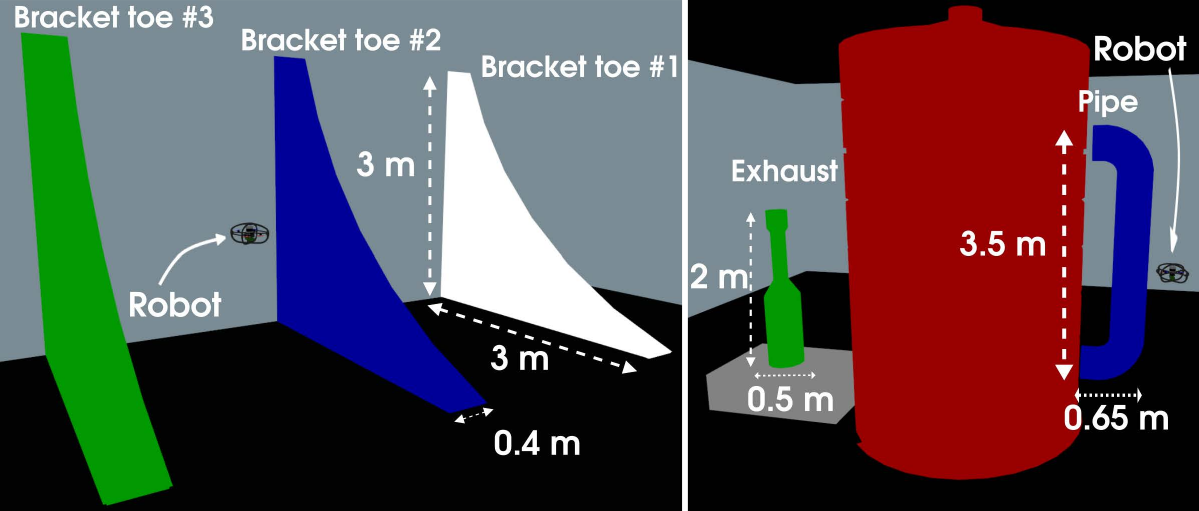}
    \vspace{-2ex}
    \caption{\textit{Left:} \ac{wbt} with three concave-shaped bracket toes. \textit{Right:} Chemical plant with three objects: pipe, tank and exhaust. For each environment once the inspection time of the semantic is passed, the \ac{rl} framework switches the semantic label to the next object of interest.}
    \label{fig:simGazebo}
    \vspace{-2ex}
\end{figure}

To deeply evaluate our approach, we present a simulation study to compare the proposed \ac{rl} planner with state-of-the-art methods for robotic visual inspection in environments containing semantic objects of interest and obstacles. The comparison is performed in two different environments: a) a \ac{wbt} with dimensions of $10 \times 10 \times 10$~\SI{}{\meter}, where the goal is to inspect three bracket toes located at the bottom of the compartment with environmental walls as obstacles and b) a simple model of a chemical plant of size $10 \times 7.5 \times 5.5$~\SI{}{\meter}, where the objective is to inspect the vertical pipe and exhaust while the tank serves as an additional obstacle. The study utilized the Gazebo simulator \cite{koenig2004design}, which implements a semantic segmentation camera model. For reference, we depict both environments in \Cref{fig:simGazebo}. To efficiently implement the local occupancy map we utilize the volumetric mapping library Voxblox~\cite{oleynikova2017voxblox}.

The performance of the proposed method is compared against a) FUEL~\cite{zhou2021fuel}, a pure exploration, and thus naturally less efficient, method, that employs a fast frontier-based exploration strategy and the sensor parameters are constrained to ensure that the planner sees all surfaces for complete coverage, b) GVI~\cite{dharmadhikari2023autonomous}, 
a method designed for volumetric exploration and simultaneous visual inspection, originally focused on ballast tanks. It performs inspection of all mapped surfaces from an optimal inspection distance, not limited to any specific type of structure
as well as c) SWAP~\cite{dharmadhikari2023semantics}, a semantics-aware exploration and inspection path planning framework. SWAP explicitly integrates semantics into the planning process, allowing the planner to volumetrically explore, generate complete mesh reconstructions of semantics, and inspect surface faces while considering image quality metrics. 

\begin{figure}[t]
    \centering
    \includegraphics[width=0.98\columnwidth]{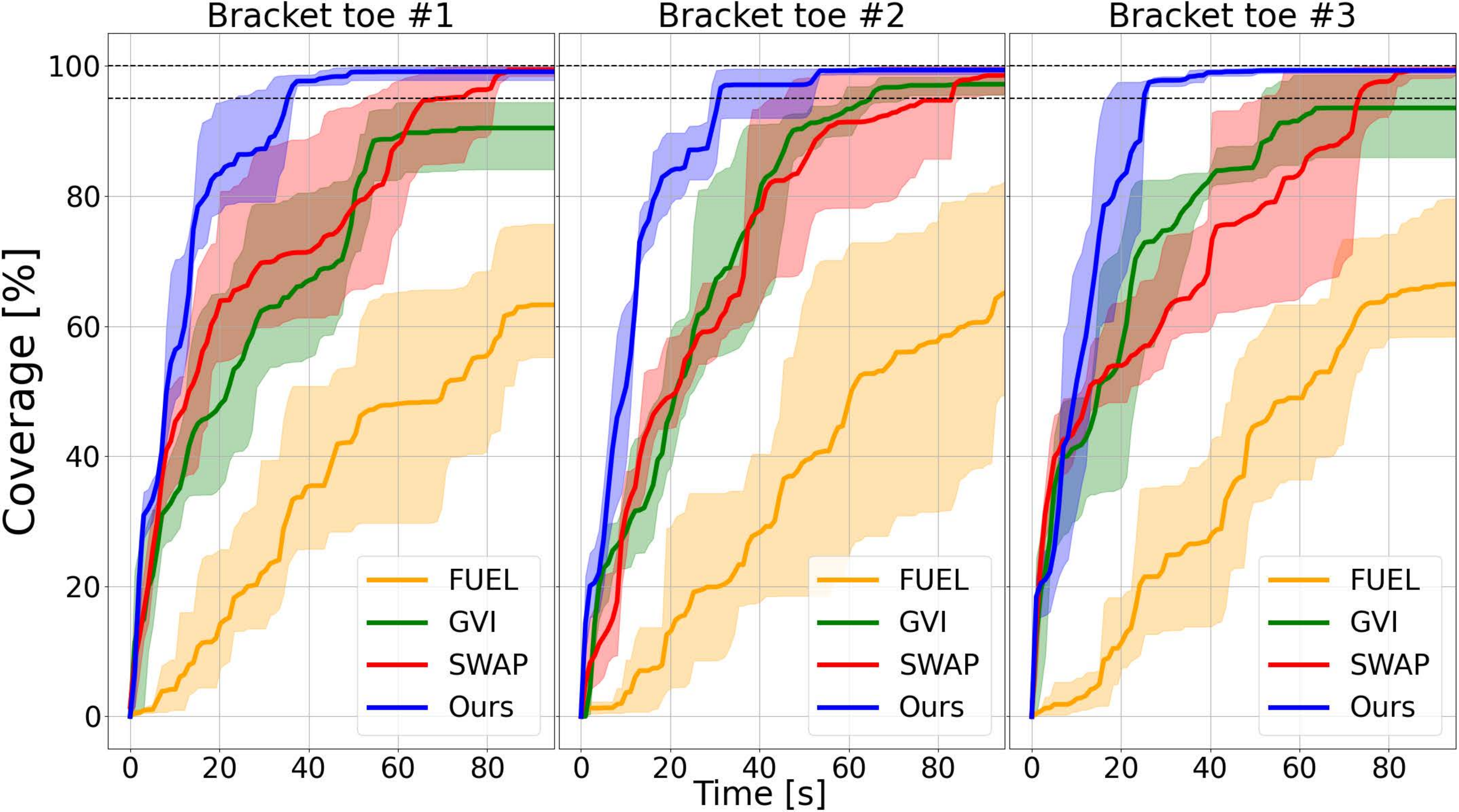}
    \includegraphics[width=0.98\columnwidth]{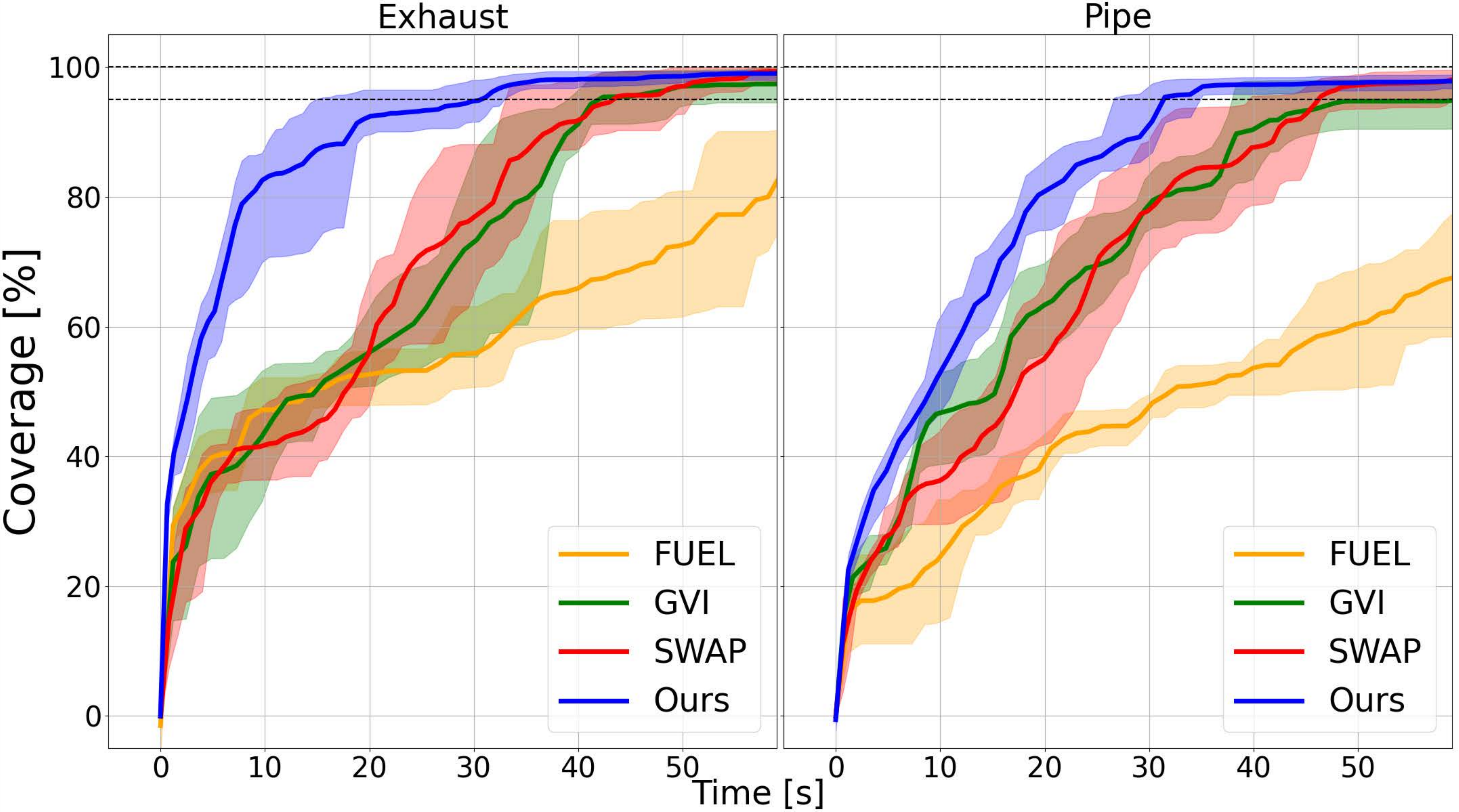}
    \vspace{-2ex}
    \caption{The percentage of the cumulative surface of each semantic observed by the camera sensor over time, considering visibility under distance and resolution constraints, by all methods. The transparent region represents the~5th–95th percentile range of surface coverage variability across runs, while the solid line indicates the mean coverage. With two horizontal lines we depict the $95\%$ and $100\%$ of the ideal feasible coverage.}
    \label{fig:simPlots}
    \vspace{-4ex}
\end{figure}
For each method, we run five visual inspection missions per environment, with consistent aerial robot starting positions across all trials. To ensure fair comparison across all methods, $\mathbf{w}_{max}$ is set to \SI{1}{\meter/\s} for linear velocity and \SI{1}{\radian/\s} for yaw-rate. To evaluate FUEL, which is not specifically designed for inspection tasks, we constrain the exploration volume to a 3D bounding box enclosing the semantic object with a \SI{0.5}{\meter} margin in all directions. GVI is modified so that its information gain metric considers only the volumetric map regions corresponding to the semantic object. The parameters of GVI and SWAP are fine-tuned to achieve the best possible performance for each method in each evaluation environment, whereas the \ac{rl} policy does not require any environment-specific fine-tuning. The parameter tuning is conducted in consultation with the authors of the respective methods and optimized to achieve greater than $95\%$ coverage while minimizing inspection time. 
To define fully autonomous inspection for the \ac{rl} method, we introduce the inspection time limit for each semantic object based on its size. Thus, for the chemical plant semantics, the maximum inspection time is set to \SI{50}{\s}, while for each bracket toe inside the \ac{wbt}, it is set to \SI{65}{\s}. The inspection timer starts when the first semantic pixel in the camera image enters the inspection distance. Once the time expires, the semantic label automatically switches to the next object. In the chemical plant scenario, the label is initially set to the pipe and then changes to the exhaust. The inspection order in the \ac{wbt} is defined from the rightmost to the leftmost bracket toe, as shown in~\Cref{fig:simGazebo}. All the semantic objects used for evaluation were not included in training and are more complex than the primitive shapes employed during training. In \Cref{fig:simPlots}, we present the average surface coverage over time, measuring the percentage of all feasible faces of the semantic mesh seen by the camera within the required inspection distance, $d_{ref}~\in~[0.8, 1.2]$~\SI{}{\meter}. The ideal feasible coverage for each semantic accounts for occlusions from nearby obstacles or parts of the target object. The surface areas of the semantics are as follows: \SI{7.85}{\meter^2} per bracket toe, \SI{2.33}{\meter^2} for the exhaust, and \SI{4.51}{\meter^2} for the pipe.

Thanks to its purely semantic-oriented exploration and inspection behavior, the proposed \ac{rl} method outperforms other inspection methods in terms of the time required for visual inspection while maintaining consistent surface coverage across runs. Unlike GVI and SWAP, the proposed method leverages agile maneuvering without following a waypoint-based trajectory. Moreover, the \ac{rl} method showcases its generalizability by successfully inspecting novel concave structures, such as the bracket toe or blue pipe, which feature significantly greater complexity than the primitive shapes used during training. Even FUEL, despite operating within a restricted exploration volume optimized for the inspection task, fails to capture all surfaces of the semantic object at the desired quality within a timeframe comparable to that of the proposed approach.

\begin{figure}[t]
    \centering
    \includegraphics[width=0.98\columnwidth]{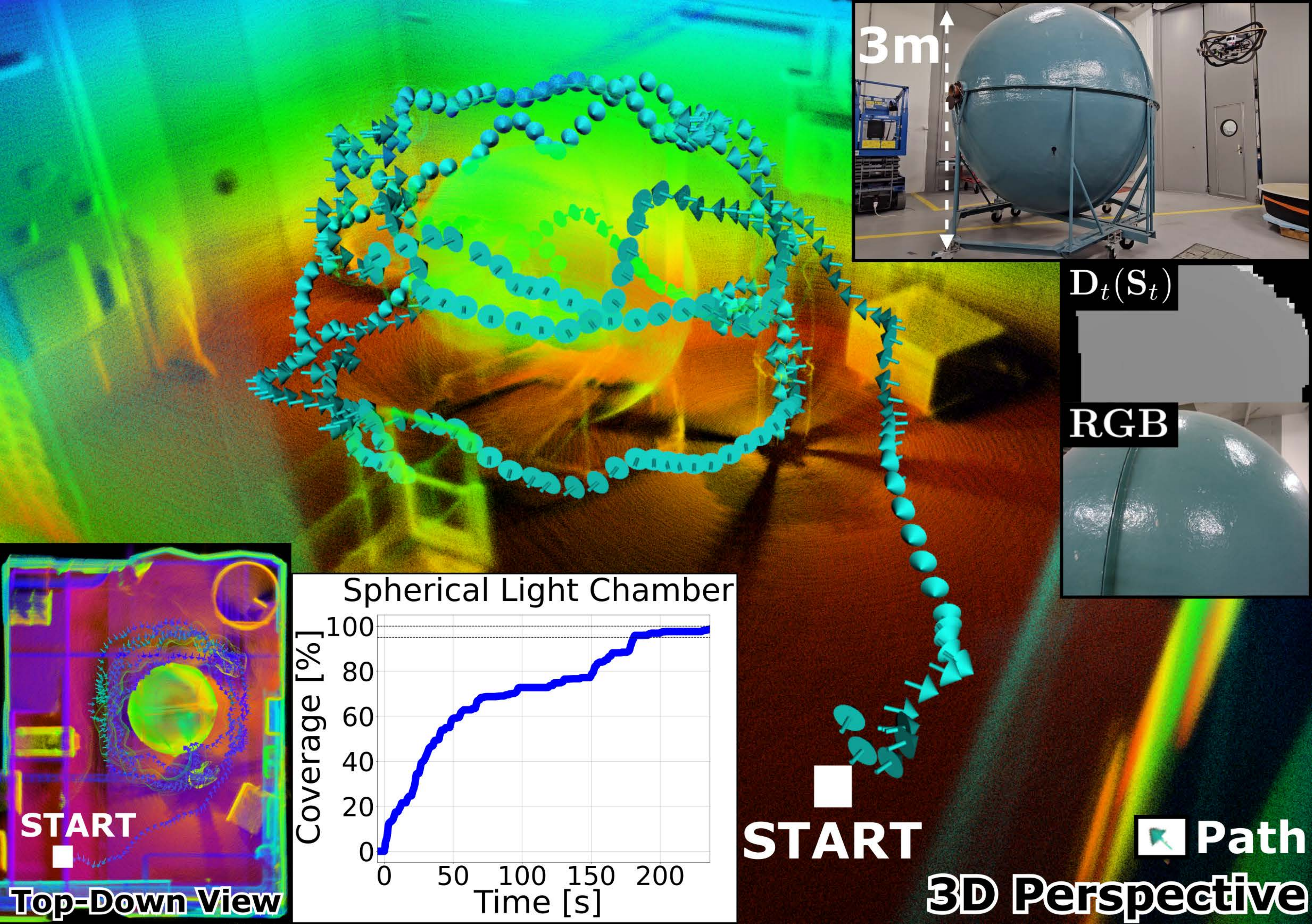}
    \vspace{-2ex}
    \caption{Single semantic inspection. The agent searches the environment to find the object of interest, performs the visual inspection, orbiting around the object while remaining collision-free and maximizing the surface coverage. The cyan arrows indicate the collision-free coverage path~$\mathcal{P}$. \textit{Left:} Surface coverage plot, \textit{Right:} Semantically masked depth image as seen by the network and corresponding RGB image.}
    \label{fig:singleSemantic}
    \vspace{-3ex}
\end{figure}

\subsection{Experimental Evaluations}
\begin{figure*}[t]
    \centering
    \includegraphics[width=0.98\textwidth]{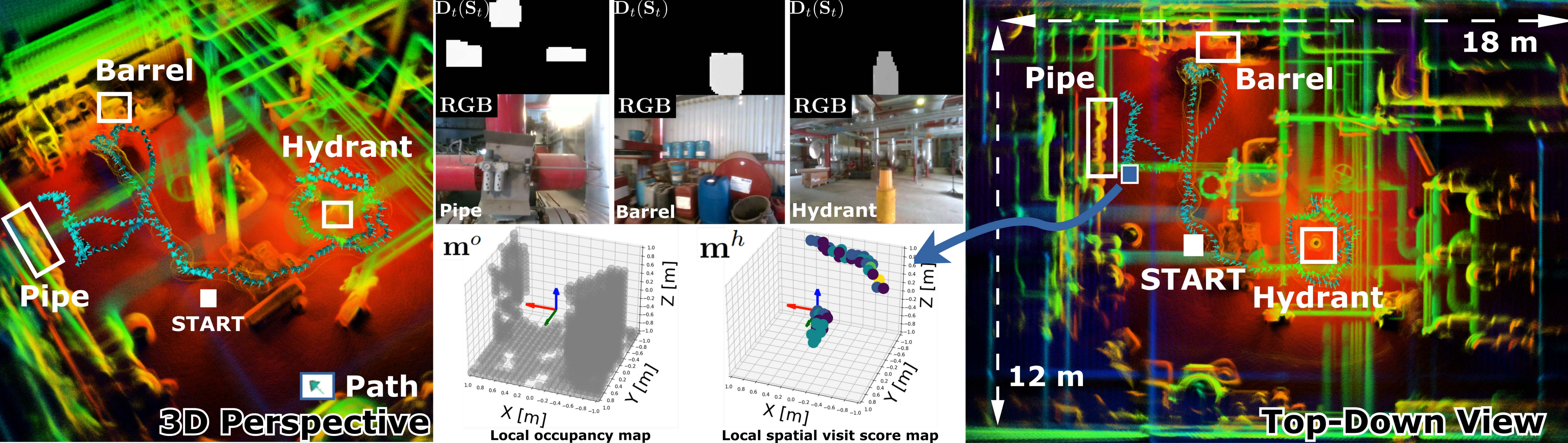}
    \vspace{-2ex}
    \caption{\textit{Left:} 3D perspective of the inspection mission with the path indicated by cyan arrows. \textit{Top row:} Semantic depth masks of the objects as observed by the network and the corresponding RGB images from onboard camera captured during the mission. \textit{Bottom row:} Instantaneous local occupancy and \ac{svs} maps observed by the robot at the location marked with blue rectangle in the map. The frame in local maps indicates the agent. \textit{Right:} Top-down view of the mission with the environment dimension indicated by white arrows.}
    \label{fig:multipleSemantic}
    \vspace{-3ex}
\end{figure*}
\begin{figure}[t]
    \centering
    \includegraphics[width=0.98\columnwidth]{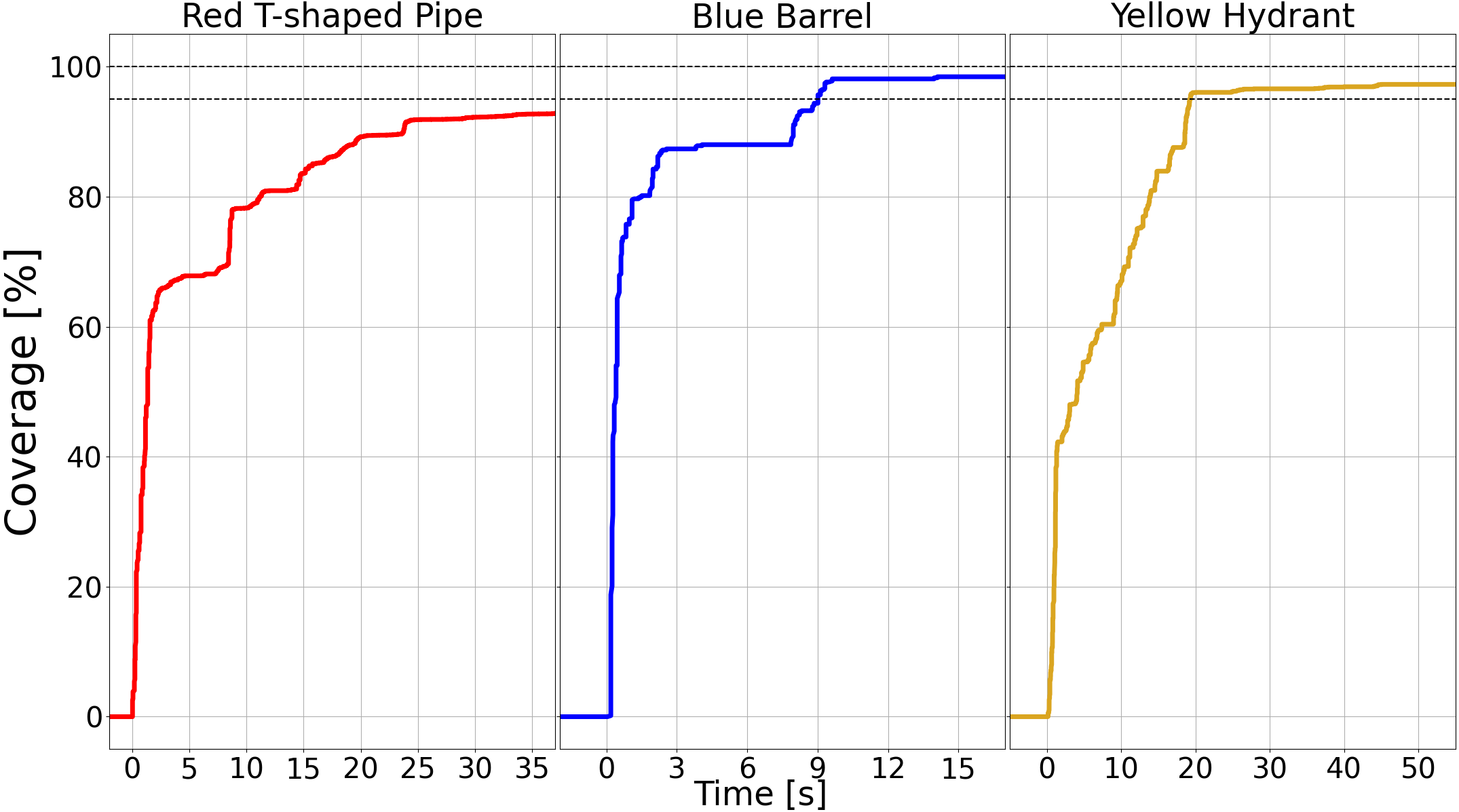}
    \vspace{-2ex}
    \caption{Surface coverage for each semantic is plotted against time, with dotted horizontal lines indicating a $5\%$ range from ideal feasible coverage.
    }
    \label{fig:surface_coverage_multiple}
    \vspace{-3ex}
\end{figure}
\label{sec:experimental}
Next, we present a set of experimental deployments of the agent onboard a quadrotor performing autonomous inspection, initially in a scene with a single semantic and then in a more complex scenario involving multiple semantic objects. Most importantly, both the semantic objects of the real-world experiments as well as the obstacles were not modeled in the training process in order to showcase the generalizability of the approach and the ability of the \ac{rl} policy to perform the inspection of unknown objects. Unlike in training and the ablation studies conducted in simulation, where the segmentation image is provided by the simulator~\cite{kulkarni2025aerialgym_journal}, in real-world experiments, we obtain the segmentation mask using the onboard RGB stereo camera Intel RealSense D455, which is mounted on the aerial robot. The technical details of the platform can be found in~\cite{harms2024neural}. D455 is set to output the RGB image with a resolution of $640\times480$. 
We deploy the masking method which involves creating a binary mask based on specific colour ranges in the HSV colour space. This technique is straightforward and efficient for isolating colours within a defined range. To reduce the motion blur and guarantee good visual inspection properties of the RGB image, we set $\mathbf{w}_{max}$ to \SI{0.3}{\meter/\s} for linear velocity and \SI{0.6}{\radian/\s} for yaw-rate. While these speed constraints increase the total inspection time, they do not compromise complete surface coverage performance and ensure robust semantic masking.
All derivations, from LiDAR-based localization and mapping exploiting the integrated OUSTER OS$0$-$64$ LiDAR to the execution of the proposed \ac{rl} agent, are running onboard within the platform's NVIDIA Jetson Orin NX compute board, with sensor observations synchronized in the control loop alongside the \ac{rl} network. The proposed method achieves an average inference time of \SI{100}{\ms} on the robot without any optimizations for faster inference.





\paragraph{Single Object Visual Inspection}
\label{sec:single}
We begin our real-world validation with a focused inspection task involving a single object, namely a spherical light chamber with diameter of \SI{3}{\meter} as shown in~\Cref{fig:singleSemantic}. The experiment tests the agent's ability to detect the semantic target and systematically maximize visual coverage of its surface. The post-processing analysis of the semantic mesh confirms the thoroughness of the inspection, validating the policy's effectiveness in single-object inspection scenarios. As illustrated in the surface coverage plot in~\Cref{fig:singleSemantic}, our approach demonstrates remarkable efficiency, with the robot executing a deliberate circling maneuver around the inspection target. The plateau observed at approximately \SI{100}{\s} into the mission occurs due to a nearby obstacle and occupancy mapping inaccuracies that prompted the agent to adjust its collision-free coverage path~$\mathcal{P}$. Despite this challenge, this methodical inspection path enables the onboard camera to capture the entire semantic surface (\SI{28.27}{\meter^2}), achieving complete coverage in under four minutes.

\paragraph{Multiple Objects Visual Inspection}
\label{sec:multiple}

In our most complex experiment, we demonstrate the framework's ability to inspect multiple objects of varying shapes that were never encountered during training. Conducted in an industrial setting, specifically within the compressor shed of a three-phase flow laboratory, the mission targets three distinct semantic objects: a red T-shaped pipe, a blue barrel, and a yellow refinery hydrant, all shown in~\Cref{fig:multipleSemantic}.

The robot begins approximately~\SI{4}{\meter} away from the first semantic target, the red pipe, requiring it to explore the local environment while navigating around nearby obstacles. The system employs a time-threshold mechanism to transition between inspection targets, shifting focus from the pipe to the barrel, and finally to the hydrant in sequence.

\Cref{fig:surface_coverage_multiple} presents the inspection coverage metrics for each semantic object, derived from post-processing analysis of the reconstructed object meshes when excluding surfaces that are impossible to inspect due to occlusions or obstacles. This distinction provides a more realistic assessment of the system's true performance.

The complete mission spans \SI{157.22}{\s}, with $64\%$ of this time dedicated to active visual inspection of the semantic targets, covering \SI{3.54}{\meter^2} in total. The remaining $36\%$ is spent on semantic-oriented search in the unknown environment, as shown in~\Cref{fig:multipleSemantic}. For the T-shaped pipe, the agent does not achieve complete coverage as the upper lid remains uninspected due to its near-horizontal orientation and positioning at a height that limits visibility within the defined inspection distance. The system still achieves an average surface inspection coverage of $96.2\%$ of all feasible semantic surfaces, demonstrating thoroughness despite the complex, multi-object nature of the task and the previously unseen object geometries.

\section{Conclusion}
\label{sec:conclusion}
A deep \ac{rl}-based planner for the autonomous semantics-aware inspection in unknown environments is presented. Through end-to-end learning, the method plans efficiently to both explore the local surroundings and perform visual inspection of the objects of interest without any prior knowledge or assumptions about the environment and the semantics. Importantly, the method assumes access only to local information about the scene. Trained on primitive geometric shapes as semantics, it is shown to being able to negotiate novel semantic objects at runtime. The comprehensive evaluation framework, comprising extensive simulation studies, detailed ablation analyses, and real-world experimental deployments, allows to demonstrate the performance of the method, how it robustly crosses the sim2real gap, and the applicability of \ac{rl} for such complex planning problems. The method is open-sourced in \url{https://github.com/ntnu-arl/semantic-RL-inspection}.

\bibliographystyle{IEEEtran}
\bibliography{./main}

\end{document}